\documentclass[journal]{IEEEtran}
\usepackage{cite}
\ifCLASSINFOpdf
\else
\fi
\usepackage[cmex10]{amsmath}
\usepackage{array}
\usepackage{mdwmath}
\usepackage{mdwtab}
\usepackage{graphicx}
\usepackage{euscript}
\usepackage{amsfonts}
\usepackage{bigstrut}
\usepackage{tabularx}
\usepackage{dblfloatfix}
\usepackage{cite}
\usepackage{amsmath}
\usepackage{algorithm}
\usepackage[noend]{algpseudocode}
\makeatletter
\def\BState{\State\hskip-\ALG@thistlm}
\makeatother
\usepackage{enumitem}
\usepackage{upgreek}
\usepackage{dsfont}
\usepackage{amsmath}
\usepackage{amsthm}
\usepackage{amssymb}
\usepackage{xcolor}
\usepackage{mathtools}

\DeclarePairedDelimiter\floor{\lfloor}{\rfloor}
\usepackage{hyperref}
\usepackage{epstopdf} % When using eps figures

\usepackage{bbm}

\begin{document}

\title{Machine Learning Framework for Sensing and Modeling Interference in IoT Frequency Bands}

\author{Bassel Al Homssi,~\IEEEmembership{Student Member,~IEEE}, Akram~Al-Hourani,~\IEEEmembership{Senior Member,~IEEE},\\ Zarko Krusevac~\IEEEmembership{Senior Member,~IEEE}, and Wayne S T Rowe,~\IEEEmembership{Member,~IEEE}.
	
\thanks{B. Al Homssi, A. Al-Hourani, and W. S. T. Rowe are with the School of Engineering, RMIT University, Melbourne, Australia. Z. Krusevac is with the Australian Communication and Media Authority. E-mail: bhomssi@ieee.org, akram.hourani@rmit.edu.au.}
\thanks{This research has been partially funded by the Smart Cities and Suburbs Program, Northern Melbourne Smart Cities Network Enabling Data to Drive Change.}
}

\maketitle

\begin{abstract}
Spectrum scarcity has surfaced as a prominent concern in wireless radio communications with the emergence of new technologies over the past few years. As a result, there is growing need for better understanding of the spectrum occupancy with newly emerging access technologies supporting the Internet of Things. In this paper, we present a framework to capture and model the traffic behavior of short-time spectrum occupancy for IoT applications in the shared bands to determine the existing interference. The proposed capturing method utilizes a software defined radio to monitor the short bursts of IoT transmissions by capturing the time series data which is converted to power spectral density to extract the observed occupancy. Furthermore, we propose the use of an unsupervised machine learning technique to enhance conventionally implemented energy detection methods. Our experimental results show that the temporal and frequency behavior of the spectrum can be well-captured using the combination of two models, namely, semi-Markov chains and a Poisson-distribution arrival rate. We conduct an extensive measurement campaign in different urban environments and incorporate the spatial effect on the IoT shared spectrum.
\end{abstract}


\IEEEpeerreviewmaketitle 

\section{Introduction}
\IEEEPARstart{T}{he} Internet-of-Things (IoT) gained an expanding momentum over the past few years serving many emerging business applications relying on wireless sensor networks. These applications range from utility metering, precision agriculture, to hazard prevention~\cite{HussainFatima2017IoTB}. Projections estimate that vast numbers of IoT devices will join the wireless network infrastructure by 2022~\cite{mobility_report}. Despite many benefits that IoT offers, challenges emerge~\cite{7815384}; a prominent one is the availability of the spectrum where IoT services are likely to underlay the current wireless network with the risk of congesting the already busy bands and increasing the existing overall interference~\cite{8030482}.%,7546934}.

Currently, spectrum authorities have allocated two main spectrum types to IoT wireless access systems; (i) a \emph{licensed} spectrum exclusive to service providers and (ii) an \emph{unlicensed}\footnote{The unlicensed spectrum in Australia is referred to as class-licensed due to sharing-focused general authorizations for spectrum use.} spectrum subject to minimal license administration for any vendor or user~\cite{ACMA_LIPD}. Many proprietary technologies and standards have emerged to cater for the specific needs of Low-Power-Wide-Area Network (LPWAN) such as  NB-IoT%~\cite{8120239}
, LoRaWAN%~\cite{specification2018lora}
, Sigfox%~\cite{specification2017sigfox}
, and Wi-SUN%~\cite{WiSUN}
~\cite{7815384}. These technologies largely differ in their approach to spectrum access; for instance, NB-IoT is widely adopted by cellular operators where typically the end user has to pay the operator a subscription fee~\cite{inproceedings}. On the other hand, technologies such as LoRaWAN, SigFox, and Wi-SUN operate in the unlicensed spectrum such as the ISM-band requiring no mandatory fees similar to conventional Wi-Fi~\cite{IoT_ACMA}. Technologies operating in the Industrial, Scientific and Medical (ISM) band which is a part of the unlicensed spectrum are exposed to interference generated not only by other IoT networks but also by other technologies that share the same frequency band~\cite{7815384,8030482}. 

Spectrum authorities impose certain regulations on the devices operating in such shared spectra in order to harmonize the co-existence of different wireless services by capping the transmit power spectral density, duty cycle, and by setting a minimum number of frequency hops~\cite{ACMA_LIPD}. These measures, however, do not completely resolve the co-existence challenge and IoT technologies still need to implement their own intelligent interference mitigation techniques. Ideally, IoT devices should be able to automatically adjust their transmit power, throughput, and bandwidth to adapt in the congested spectrum while maintaining an acceptable packet success rate~\cite{8935360,8631733}.

To understand the traffic behavior in a given frequency range, typically spectrum analyzers (SAs) are employed to capture the power spectral density in the channel with respect to time~\cite{7460899,6933906}. SAs are usually operated in the \emph{frequency-sweep} mode where the capturing occurs in sweeps across the measurement range. The measurement range is thus divided into smaller bins that are visited periodically by the SA~\cite{8403749,4213046}. However, most IoT transmissions have channel activities limited to the microsecond scale, hence, methods involving frequency sweeping are limited in time resolution inherently leading to higher misdetection.

To assess the performance of the network, typical practical approaches incorporate the interference as a single value; the average measured interference in the channel. However, even in channels with high average interference, the channel is not constantly busy and has opportunities that are available. In this paper, we propose a framework that enables IoT transmissions' autonomous sensing and detection. The framework utilizes the robustness of software defined radios (SDRs) to capture the IoT spectrum enabling higher time resolution and more accurate representation of the frames in the spectrum. The framework reduces the uncertainties generated from conventional energy detection methods~\cite{6933906} using a machine learning. In addition, we model the temporal and frequency correlation of IoT frames by utilizing the proposed framework on real-measurements in different urban environments. In this respect, our contributions are summarized as follows:
\begin{figure}
	\normalsize
	\centering
	\includegraphics[width=0.9\linewidth]{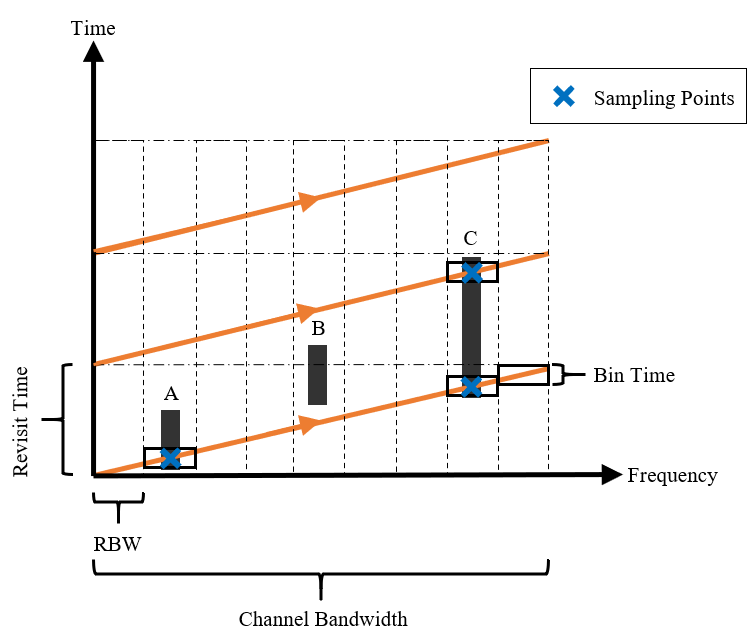}
	\caption{Illustration of SA operating in the frequency-sweep mode, the solid straight line demonstrates the sweep of the SA across frequency. Sampling occurs where the SA sweep intersects with a transmission. The SA calculates the average, positive peak, or negative peak in each time bin for every RBW.}
	\label{Fig_AvgPower}
\end{figure}
\begin{itemize}
	\item Novel technology agnostic machine learning framework capable of autonomously identifying the location and extracting the statistics of the transmissions in the unlicensed frequency bands.
	\item Unsupervised machine learning application for the reduction of the probability of false alarm and misdetection present in conventional energy detection.
	\item Modeling of IoT shared spectrum temporal correlation utilizing continuous-time Markov chain.
	\item Application of a Poisson arrival rate model to capture the behavior of the present interference traffic in the channel in different spatial environments.
\end{itemize}
This research stemmed from work related to the Northern Melbourne Smart Cities Network project~\cite{NMSCP} for the deployment of a large-scale LoRaWAN network in Victoria, Australia.
\section{Related Work}\label{Sec_Related}
A wireless radio channel of interest is subject to interference from devices sharing the same band. To capture the nature of this interference, occupancy studies utilizing spectrum sensing techniques are implemented, many of which are summarized in~\cite{6933906}. The authors detail existing studies that mainly utilize conventional swept-tuned SAs to monitor the spectrum followed by energy detection to deduce the spectrum occupancy. The SA is traditionally operated in the frequency-sweep mode~\cite{7563317} where the channel bandwidth is divided into smaller partitions (bins) with a width referred to as the resolution bandwidth (RBW). The SA sweeps across all the bins before returning to the first bin and starting over again. Each sweep is not instantaneous and requires a certain amount of time to complete called the \emph{revisit time}~\cite{ITU}. However, IoT transmissions (frames) usually have short Time-on-Air (ToA) when compared to the revisit time, hence an SA might not be capable to accurately capture the frames' behavior. Depending on the SA configuration, the power reported from each bin is calculated using different methods; the \emph{average}, the \emph{positive peak}, or the \emph{negative peak} of the power across the bin~\cite{SpectrumAnalysisBasics}. If the ToA is smaller than the bin time while the SA is utilizing the average configuration, the \emph{average power} may decrease in magnitude leading to misdetection. If the SA is utilizing the \emph{positive peak} configuration, the SA is capable of successfully reporting the power of the short transmission, however it will automatically assume that its ToA is equivalent to the revisit time when in fact it might be smaller. Similarly, the \emph{negative peak} configuration is capable of capturing the transmission but similar to the previous configuration overestimates its actual ToA. Fig.~\ref{Fig_AvgPower} illustrates the SA frequency-sweep mode reporting three different situations that the SA may encounter in the spectrum. (i) the SA reports that the channel is occupied for the entirety of the revisit time while the actual ToA for frame A is significantly shorter, (ii) frame B is misdetected due to its short ToA and misalignment with the sampling points of the SA, and (iii) the SA reports higher occupancy due to frame C's short ToA similar to frame A. SA sensing for short bursts of IoT signals may overestimate the occupancy such as the cases for frames A and C or vice-versa for frame B.

Spectrum occupancy is typically reported using the duty cycle measure which corresponds to the probability that the received power in a given channel is above a certain threshold~\cite{8115264,8691447}. However, the duty cycle is highly dependent on the measurement environment and may suffer from false alarm and misdetection uncertainties when obtained by energy detection method~\cite{4840476}. Other works have reported the occupancy in terms of the captured channel power probability distributions such as the cumulative distribution function (CDF) and probability distribution function (PDF) such as in~\cite{4213046} where the authors used amplitude probability distribution (APD) to measure the spectrum sparsity in the frequency bands. While APD models are less affected by the aforementioned uncertainties, they do not provide a temporal correlation of the occupancy in the channel. To model the temporal behavior, some work have explored utilizing many variations of Markov and semi-Markov chains such as in~\cite{250336,5936273,5506438}. These models are used to extract the average dwell time modeled using first order Markov models. The average dwell time represents the average time that the channel remains occupied before it can be free again. The accuracy of those models, nevertheless, is highly dependent on the false alarm and the misdetection uncertainties~\cite{7460899}. Models such as Hidden Markov Models (HMM) have been proposed~\cite{6918501}. However, while HMM models can be used to exploit the uncertainties and reduce their effects, they are only discrete in nature and do not fully capture the continuous temporal variations in the channel.

A more recent survey conducted on measurement campaigns and interference maps in~\cite{7460899} concludes that the choice of the measurement campaign parameters such as bandwidth, antenna location, and total measurement time should depend on the technology and radio channel behavior. Moreover, deep learning has been examined on the detection and classification of transmissions in a channel~\cite{8267032}. Such methods methods assume that the receiver has prior knowledge of the signal location on the frequency-time plane. In order to encounter this constraint, manual labeling is utilized to identify the location and statistics of the frames in the spectrum prior to the application of the machine learning classifier~\cite{8977513}.	Recall that the ISM-band is unlicensed, hence is unrestricted to any technology~\cite{8594702} making methods such as manual labeling inadequate to implement. Therefore, an autonomous technology agnostic framework is required in order to enable the capture of the IoT transmissions in the unlicensed spectrum to facilitate classification.

In this paper, we propose a framework capable of autonomously capturing IoT signals without the need for prior knowledge of their locations. The framework proposed in this paper relies on using an SDR coupled with energy detection to obtain the observed occupancy. To improve energy detection, an unsupervised machine learning algorithm is applied to autonomously cluster, filter, and reinforce the captured transmissions in various urban environments. The machine learning algorithm reduces the false alarm and misdetection uncertainties preserving the continuous temporal variations present in the channel. The framework obtains the statistics in terms of bandwidth and time-on-air without the prior knowledge of the transmissions' modulation scheme. To the authors’ knowledge, the framework presented in this paper has not been previously addressed in the literature.

\section{Energy Detection Model}\label{Sec_Occupancy}
A transmission of an IoT frame between a transmitter and a gateway over a radio channel may experience collisions that lead to errors or in some cases complete failure. The collisions mainly occur due to interference generated from other transmissions in the channel. In this case, the radio channel is considered to be in the \emph{busy} state. On the other hand, if the transmission does not experience interference, the radio channel is considered to be in the \emph{idle} state. The radio channel occupancy space comprises of two states; ${\mathbb{S} = \{S_0,S_1\}}$ where $S_0$ is the idle state and $S_1$ is the busy state. In order to observe the occupancy state of the channel, a monitoring device such as an SDR is utilized. However, due to uncertainties that typically stem from the device's thermal noise, the channel may be observed busy when in fact is actually in the idle state, and vice-versa. As a result, we define the observed channel occupancy space as ${\mathbb{H} = \{H_0,H_1\}}$ where $H_0$ is the \emph{observed idle} state and $H_1$ is the \emph{observed busy} state. The hidden and observed channel occupancy states are modeled as follows,
\begin{equation}
\resizebox{0.91\hsize}{!}{%
\begin{array}{l l l}
\begin{array}{l}
\mathbb{S} = 
\begin{cases}
S_0: & \text{Idle}\\
S_1: & \text{Busy}\\
\end{cases}
\end{array}
,
&
\text{and}
&
\begin{array}{l}
 \mathbb{H} = 
\begin{cases}
H_0: & \text{Observed Idle}\\
H_1: & \text{Observed Busy}\\
\end{cases}
\end{array}
.
\end{array}
}
\end{equation}
Various detection methods responsible for generating the spectral occupancy from the captured power spectral density are well established in the literature and choosing the method to implement depends on the particular technology~\cite{6933906}. 

In this paper, we deploy the energy detection method so that we do not restrict our work to a particular technology. Energy detection is a hard decision-making technique which assigns the state of the channel purely on the threshold power. Therefore, the threshold power $\theta$ is the deciding factor; if the received power is higher than the threshold, the channel is observed as busy, whereas observed as idle if otherwise. The observed occupancy of the spectrum is therefore modeled as follows,
\begin{equation}\label{Occ_Matrix}
B[n,k] =
\begin{cases}
1       & \quad \text{if } X[n,k]  \geq \theta\\
0  		& \quad \text{if } X[n,k] < \theta
\end{cases},
\end{equation}
where $X[n,k]$ is the discrete captured signal power spectral density, $n$ is the discrete time variable and $k$ is the discrete frequency variable.

Two distinct uncertainties occur and their effect is quantified by their probabilities.
\begin{figure}
	\normalsize
	\centering
	\includegraphics[width=\linewidth]{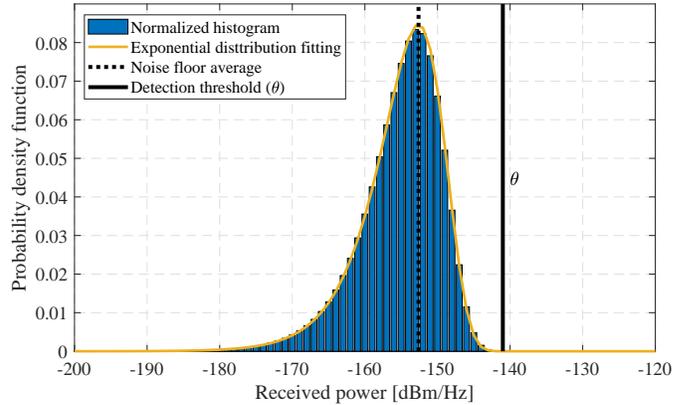}
	\caption{PDF of measured noise with an average noise floor $-154$~dBm/Hz captured with matched load termination. The threshold level obtained is $\theta=-140$ dBm/Hz for a selected probability of false alarm $p_\text{f}=10^{-6}$. The example parameters are explained further in Table~\ref{Table_Setup}.}
	\label{Fig_Faraday}
\end{figure}
The probability of a channel  to be observed busy when it is in the idle state is defined as the probability of false alarm ${\mathbb{P}(H_1|S_0)=p_\text{f}}$ and occurs when the noise power spikes higher than the threshold level. On the other hand, the probability of the channel to be observed idle  when it is in the busy state is defined as the probability of misdetection ${\mathbb{P}(H_0|S_1)=p_\text{m}}$ and occurs when a transmission's power level is lower than the threshold. Furthermore, the threshold selection requires the knowledge of the statistical distribution of the monitoring device noise and the transmissions' power levels. The SDR noise power levels are captured by replacing the antenna with a matched load to eliminate any interfering signals~\cite{8403749}.  By generating the CDF empirically through the Kaplan-Meier method~\cite{294997}, the threshold level can be obtained by setting a fixed probability of false alarm as follows,
\begin{equation}
\begin{array}{l}
p_\text{f} =  \mathbb{P}(H_1|S_0) = \mathbb{P}(X>\theta) =  1 - F_\text{N}(\theta), 
\end{array}
\end{equation}
where ${F_\text{N} = 1 - \exp(-\theta/\sigma^2)}$ is the CDF of the noise power for an AWGN channel with average noise power $\sigma^2$ and the threshold is in linear form. Fig.~\ref{Fig_Faraday} illustrates the PDF of the captured noise power. Since the bandwidth is relatively small, the noise power spectral density can be considered constant across for the measurement frequencies. 
\begin{figure*}
	\normalsize
	\centering
	\includegraphics[width=0.9\linewidth]{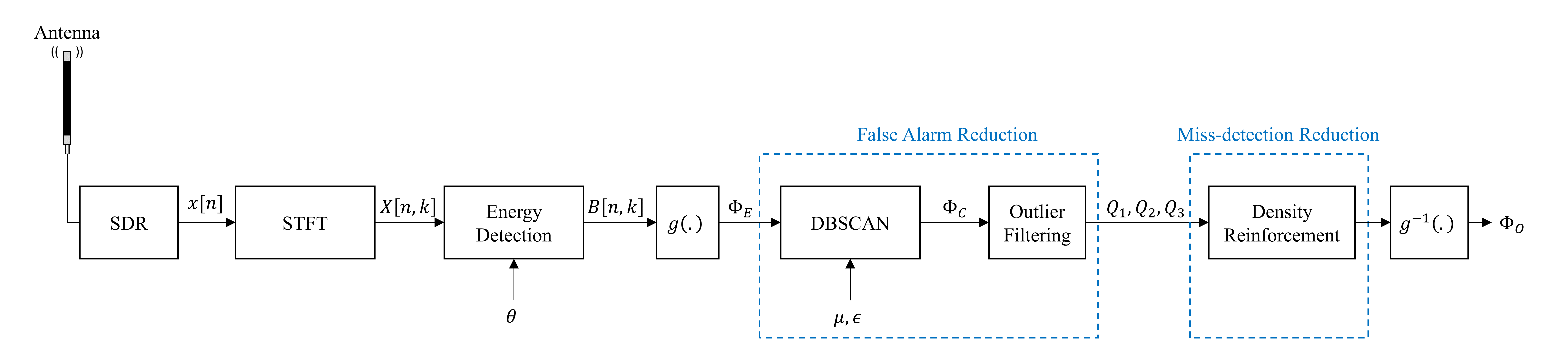}
	\caption{Block diagram of the proposed machine learning algorithm in reducing the false alarm and misdetection uncertainties.}
	\label{Fig_FlowChart}
\end{figure*}

The ISM-band contains interference traffic from many different technologies with various data rates and transmit powers. these parameters are unknown to the monitoring device, therefore, difficulty arises in estimating the probability of misdetection deeming models such as HMM~\cite{6918501} unfeasible. Thus, we propose an algorithm that autonomously identifies and improves the transmitted frames obtained from energy detection, limiting the uncertainties arising from false alarm and misdetection.
\section{Machine Learning for Detection Improvement}\label{Sec_Cluster}
IoT data is usually transmitted in the form of frames that span over a certain ToA and bandwidth. Thus, using high spectrogram resolution, most IoT frames are expected to appear with sharp edged rectangle of points ``near'' each other on the time-frequency plane. The detection of an IoT frame will appear as closely correlated points due to the discrete nature of the spectrogram matrix, whereas falsely detected noise is expected to appear as scattered points with no dominant structure. The observed occupancy matrix obtained in (\ref{Occ_Matrix}) is a binary matrix where a cell with the value ``1'' represents an observed busy state. On the time-frequency plane, the observed occupancy point process is the union of the points as follows,
\begin{equation}
\Phi_{B} =  \{\mathds{1}_{x\geq\theta} : x\in X \},
\end{equation}
where $x$ is the spectrogram matrix.  Fig.~\ref{Fig_FlowChart} illustrates a block diagram of the proposed framework to enhance the energy detection. We define a frame as a cluster of points in the shape of a \emph{rectangle} with a height and width corresponding to the ToA and bandwidth respectively as illustrated in the bottom left sub-figure in Fig.~\ref{Fig_Points}. Hence, points that form a cluster are considered a single transmission frame in comparison to those that are irregularly scattered points forming no specific shape and are therefore considered falsely detected noise.

\subsection{Machine Learning Clustering}
In order to distinguish the transmission frames from falsely detected noise points, we require a method able to identify the clusters with no previous knowledge of the numbers and sizes of frames while neglecting irregular points deeming supervised clustering technique such as k-means unsuitable. A suitable clustering algorithm for such applications is the Density-Based Spatial Clustering of Applications with Noise (DBSCAN)~\cite{DBSCANCite}. DBSCAN is dependent on the Euclidean distance between the points. However, the time-frequency plane is composed of two distinct domain units; where time is measured in seconds and frequency is measured in Hertz. A typical IoT frame has a ToA in the microseconds and a bandwidth in the kilohertz range. On the contrary, the axes of the Euclidean plane are both of the same unit length. As a result, the observed occupancy point process needs to be mapped from the time-frequency plane $\mathbb{R}_{tf}^2$ to the Euclidean plane $\mathbb{R}^2$ using a transformation function, we denote as $g(.)$ prior to implementing the DBSCAN. For IoT applications in specific, we introduce a mapping function that relies on a scaling factor $\delta$ to control the increment in the frequency step ${0<\delta<1}$ and define the mapped observed occupancy point process $\Phi_{E}$ as follows,
\begin{equation}
\Phi_{E} = g(\Phi_{B}) \triangleq \bigcup_{x\in\Phi_{B}}\{g(x) \},
\end{equation}
where ${\Phi_{E} \in \mathbb{R}^2}$, and the mapping function ${g:(\mathbb{R}_{tf}^2)\rightarrow\mathbb{R}^2}$ for a point located at $x$ on the time-frequency plane is defined as follows,
\begin{equation}
g(x) = 
\left[\begin{array}{c c}
x_t, & \delta x_f
\end{array}\right],
\end{equation}
where $(x_t,x_f)$ are the time and frequency coordinates of $x$. The value is dependent on the resolution of the system and the desired distance scaling in the time-frequency plane. In our algorithm ${\delta = 0.5}$ was chosen for a measurement resolution of [0.5 msec, 20 kHz], since most narrow-band IoT communications are on the scale of hundreds of kHz whereas much smaller in time domain.
%A good description of the DBSCAN algorithm can be found in
%and for completeness we indicate below the basic concept of cluster formation in DBSCAN. DBSCAN chooses a random point and scans around it at a radius $\epsilon$ looking for other points. All the points found are considered that point's neighbors. If no neighbors are found, the point is considered a random point and is labeled as noise, in this case the algorithm moves on to the next unlabeled point. However, if neighbor points are found, the point is considered a core cluster point and the algorithm iterates over each of the neighbor points looking for their respective neighbors adding them to the same cluster. If the number of points in the cluster is greater than $\mu$, the cluster holds and the core point and its neighbors are labeled by a cluster number. If it is below, only the core point is labeled as noise and the algorithm moves on to the next unlabeled point until all the points are labeled as either cluster points or noise.
The accuracy of the clustering in DBSCAN subsequently relies on (i) the minimum number of points in a cluster $\mu$ and (ii) the radius of scanning $\epsilon$~\cite{DBSCANCite}. We propose an \textbf{autonomous DBSCAN} method to ensure that the algorithm is completely autonomous and adaptive and determines the values of the two parameters based on the spectrum measurements solely with no manual inputs.

\subsubsection{Minimum Number of Points in a Cluster $\mu$}
We follow a heuristic method found in~\cite{suthartechnical} which relies on the logarithm of the total number of points as follows,
\begin{equation}
\mu = \floor*{\ln\left(\sum_{i=1}^{N_p}\mathds{1}_{x_i\in\Phi_{E}}\right)},
\end{equation}
where $N_p$ is the total number of points to be clustered. The reason for choosing this method is that if the total number of points increases, the minimum number of points in a cluster also increases logarithmically making the condition to form a cluster more difficult to achieve.
\subsubsection{Scanning Radius $\epsilon$}
We propose using the nearest neighbor (NN) distance method that has been reported in~\cite{Gaonkar2013AutoEpsDBSCAND}. To obtain the radius, we are interested in the distance between the points and their $\mu{\text{th}}$ NN where $\mu$ is the minimum number fo points in the cluster. Thus the $\mu$-NN distance is calculated for every point as follows,
\begin{equation}
\mu-\text{NN}(x) = ||x - x_N||,  x \in \Phi_{E},
\end{equation}
where $x_N\in\Phi_{E}$ is the $\mu^{\text{th}}$ nearest neighbor, and $\mu\text{-NN}(x)$ is the distance between $x$ and $x_N$. The sorted $\mu$-NN distance is a curve that has $\epsilon$ as its knee point.
%\begin{figure} 
%	\normalsize            
%	\centering
%	\includegraphics[width=\linewidth]{Figures/KneePt.eps}
%	\caption{Nearest neighbor distance versus the number of points. The black cross represents the obtained knee point and corresponding $\epsilon$ is the distance magnitude of the knee point.}
%	\label{Fig_knee_pt}
%\end{figure}
The knee of a curve by definition is the local maximum of a curvature and is the furthest value from a line extended from the beginning to the end of the curve~\cite{5961514}.
% as illustrated in Fig.~\ref{Fig_knee_pt}.
We define the cluster point process as the union of the clusters identified by the DBSCAN algorithm represented as follows,
\begin{equation}\label{Eq_Cluster}
\Phi_{C} = \bigcup_{j = 1}^{N}C_j= \bigcup_{j = 1}^{N}\bigcup_{x\in\Phi_{E}} x\mathds{1}_{M(x) = j} ,
\end{equation}
where $C_j$ is the $j^{th}$ cluster and $M(x)$ is a measure which represents the cluster label ${1\leq M(x)\leq N}$ where each point has a label and the points deemed not part of any cluster are donated a label of ``0'' and are dropped. $N$ is the total number of clusters identified. Note that each clusters may still have points that might be generated from falsely detected noise referred to as outliers and are filtered out in the next subsection.

\begin{figure}
	\normalsize
	\centering
	\includegraphics[width=\linewidth]{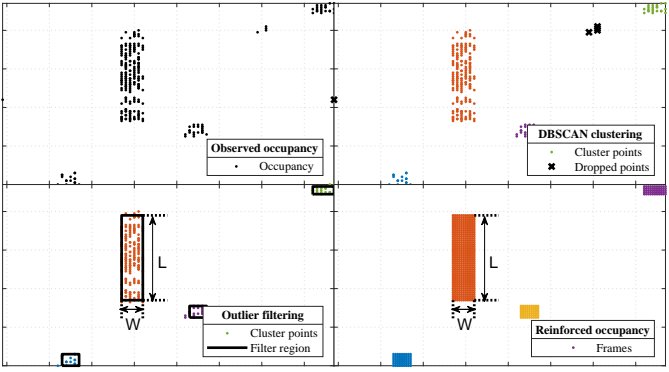}
	\caption{Spectrum snap illustrating the proposed algorithm. Conventional energy detection $\Phi_E$ (upper left). DBSCAN cluster points $\Phi_C$; each color represents a cluster and dropped noise points are represented as crosses  (upper right). Length and width identification (lower left). Outlier removal and re-enforced clusters $\Phi_O$ (lower right).}
	\label{Fig_Points}
\end{figure}

\subsection{Outlier Filtering}
The clusters obtained $\Phi_{C}$ may contain outlier points. These outliers are in essence falsely detected noise located at a distance smaller than $\epsilon$ from the closest point in a cluster. To identify the outliers, we deploy a filter based on Tukey fences. This method relies on the location statistics of the points instead of their number~\cite{Seo2006ARA}. A point is considered an outlier if it is outside a positive constant multiple $\kappa$ of the interquartile range as such, $h = \kappa[Q_3(.)-Q_1(.)]$, where $Q_1(.)$ and $Q_3(.)$ are the 25$^{th}$ and 75$^{th}$ percentiles respectively. Since the points are two dimensional in space, two filters are considered; one for each axes. Let the cluster of points be part of an ideal rectangle, then the coordinates have a uniform distribution with a pdf of ${f_R(x) = 1/K}$ where $K$ is the magnitude of the side length. In order to preserve the points inside the boundaries, we set the filter to the dimension $\kappa=2$. Therefore, the boundaries, i.e. the length and width, of the $j^{th}$ cluster are respectively obtained as follows,
\begin{equation}\label{dimensions}
\begin{array}{l}
L_j = 2\times[Q_3(x_j)-Q_1(x_j)], \\
W_j = 2\times[Q_3(y_j)-Q_1(y_j)], \\
\end{array}
\end{equation}
where $(x_j,y_j)$ are the coordinates of the points in cluster $j$. Any point that is located outside these regions is thus considered an outlier and is dropped from the cluster. Furthermore, the centroid coordinates $(x_{0_j},y_{0_j})$ are assumed to be the center of the $j^{th}$ cluster. Fig.~\ref{Fig_Points} also illustrates a closer snap at some of the frames and the framework identifying, filtering, and reinforcing the number of points for those frames. Note that in order to for this method to be reliable, high resolution in time is required to ensure that the transmissions are well-represented by many points.
\subsection{Re-enforcement and Occupancy}
Due to misdetection uncertainty, the filtered clusters may have missing points (voids) leading to a homogeneous density. However, now that we obtained the cluster boundaries from (\ref{dimensions}), we drop all remaining points and assume that all points that can exist within those boundaries are transmission points whether they are captured or misdetected. As a result, each cluster represents a well-defined rectangle with bounded dimensions with no voids. Collectively, the  mapped estimated occupancy point process is the collection of those clusters referred to as $\Phi_R$. Although the number of points are increased to fill the rectangular bounds, the total number of clusters is unchanged preserving the number of captured frames. In order to analyze the frames' ToA and bandwidth, $\Phi_R$ is returned to the time-frequency plane using the inverse mapping function and the estimated occupancy point process is obtained as follows,
\begin{equation}\label{Eq_Packets}
\Phi_{O} = g^{-1}(\Phi_{R}) \triangleq \bigcup_{x\in\Phi_{R}}\{g^{-1}(x) \} ,
\end{equation}
\begin{figure*}[b]
	\normalsize            
	\centering
	\includegraphics[width=\linewidth]{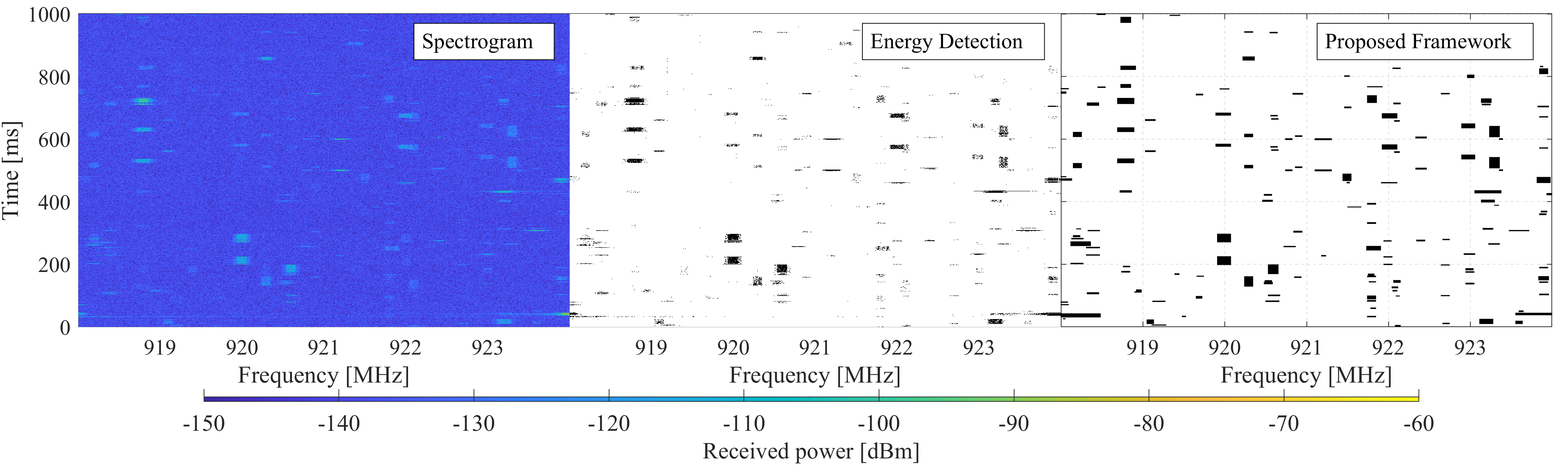}
	\caption{A snapshot of collected spectrogram in Melbourne CBD; raw captured spectral power density samples (left), occupancy from energy detection (middle), and occupancy using proposed framework $\Phi_O$ (right). The algorithm incurred a reduction of 0.5\% and 5\% on false alarm and misdetection respectively.}
	\label{Fig_Spectrum}
\end{figure*}
where , $g^{-1}(.)$ represents the inverse mapping function  ${g^{-1}:\mathbb{R}^2\rightarrow(\mathbb{R}_{tf}^2)}$ and ${\Phi_{O}\in(\mathbb{R}_{tf}^2)}$. The clustering algorithm is responsible for cleaning the noise points and reinforcing the IoT transmissions' shapes. The estimated occupancy matrix is thus obtained from the estimated occupancy point process as follows,
\begin{equation}\label{Occ_Matrix2}
O[n,k] =
\begin{cases}
1       & \quad \text{if } x  \in \Phi_{O}\\
0  		& \quad \text{if } \text{otherwise}
\end{cases} ,
\end{equation}
where $x = \{n,k\}$. The frequency bin duty cycle is calculated as follows,
\begin{equation}
\Psi[k] = \frac{1}{T_s}\sum_{n=0}^{T_s} O[n,k] ,
\end{equation}
where $T_s$ is the total time of the measurement duration in a site and is calculated as ${T_s = (N_s-1)/f_s}$. As a result, the overall duty cycle of the whole band is the average of the frequency bin duty cycle obtained as follows,
\begin{figure}
	\normalsize
	\centering
	\includegraphics[width=\linewidth]{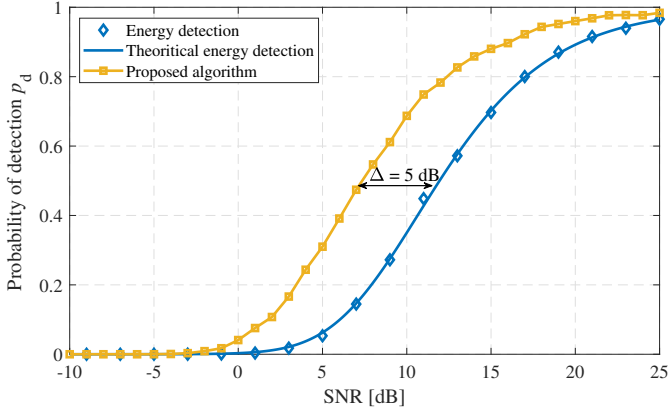}
	\caption{Probability of detection for the emulated signal having a bandwidth of $100$~kHz and a ToA of $200$~ms. We vary the SNR of the injected signal and keep the probability of false alarm constant at $p_f = 1e-6$ with $\theta = -140$ dBm/Hz.}
	\label{Fig_PoC}
\end{figure}
\begin{equation}
\Psi = \frac{1}{N_f}\sum_{k=1}^{N_f} \Psi[k] .
\end{equation}

\subsection{Improvement Analysis}
\begin{figure}
	\normalsize
	\centering
	\includegraphics[width=\linewidth]{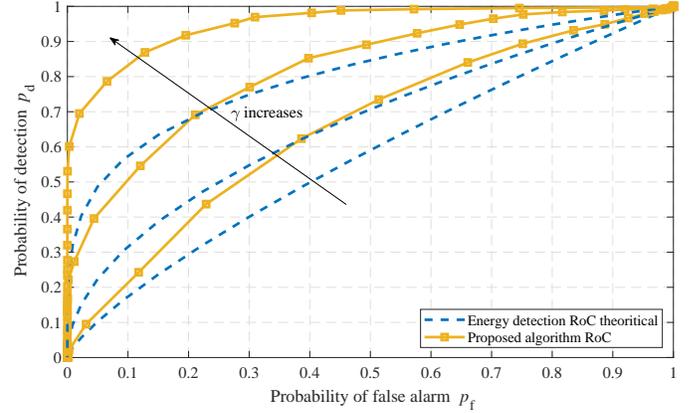}
	\caption{ROC curves for the proposed framework and energy detection for SNRs -5,0, and 5 dB while the threshold level is varied.}
	\label{Fig_RoC}
\end{figure}

In order to quantify the enhancement introduced by the proposed framework in comparison to energy detection, we inject a simulated signal with Rayleigh fading into an AWGN channel. Unlike the case with spectrum monitoring, the signal's signal-to-noise ratio $\gamma$ (SNR) and bandwidth are predefined and known prior to detection. To ensure that the AWGN channel is emulating the SDR receiver, we assign the channel and equivalent noise floor. The probability of false alarm for energy detection is the noise CCDF $p_\text{f} = \exp(-\theta/\sigma^2)$ and the probability of misdetection is $p_\text{m} = \exp(-\theta/\sigma^2)^{\frac{1}{1+\gamma}}$ for a Rayleigh fading channel with no diversity, obtained and simplified from~\cite{4063501}. By varying the SNR while maintaining a constant threshold and comparing the detected signal by the proposed framework in comparison to the injected signal, we numerically calculate the probability of detection as illustrated in Fig.~\ref{Fig_PoC}. Furthermore, we repeat the previous steps while varying the threshold and inherently the probability of false alarm while varying the SNR, we are able of numerically extracting the receiver operating characteristic (ROC) of the proposed framework. Fig.~\ref{Fig_RoC} illustrates the ROC curves for both the proposed framework and energy detection. The framework displays significant detection enhancement across all SNRs. A snapshot of applying the framework on real measurements in Site 1 is illustrated in Fig.~\ref{Fig_Spectrum}. The objective of the algorithm is to identify the location of the frames and extract the frames' statistics. Note that while the accuracy of the algorithm will fall when high transmission overlaps occur, nevertheless, it will still pick up the ``perceived" frames and identify their locations and statistics. Moreover, the occupancy obtained is for a certain threshold level and to obtain the performance of a particular communication link, the threshold can be manipulated to be equal to the sensitivity margin of the link.

\section{Spatiotemporal Interference Modeling}\label{Sec_Model}
In this section, we utilize the obtained occupancy from the proposed algorithm to predict the traffic in the spectrum for the time and frequency domains. These models enable IoT devices to predict the average busy time of the spectrum and number of other potential interfering transmissions.
\subsection{Temporal Model}
The proposed framework in the previous section estimates the channel occupancy model (interference) based on the observed occupancy states introduced previously in Section~\ref{Sec_Occupancy}. The radio channel occupancy model can now be simplified to one occupancy space state, namely $\mathbb{S}$. Let the channel state $s(t)$ be a function of time since the state is varying with time and is a stochastic process. Where $s(t) = S_0$ means that the channel is in the idle state whereas $s(t) = S_1$ mean that the channel is in the busy state. 
\begin{figure}
	\normalsize
	\centering
	\includegraphics[width=\linewidth]{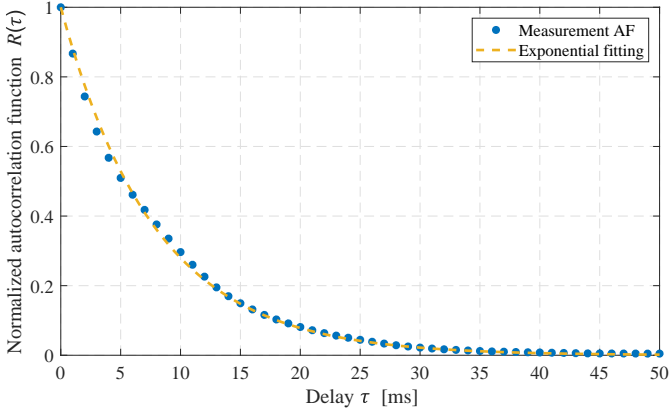}
	\caption{Comparing the collected measurements with the proposed exponential autocorrelation function $R(\uptau)$. Fitting average is ${\uptau_{\text{corr}} = 7.85}$ ms for Site~1.}
	\label{Fig_AutoCorr}
\end{figure}
In order to study the temporal correlation between two time instances, we invoke the concept of the autocorrelation function (AF). AF provides a measure of similarity between the channel state at different points in time, namely $t$ and $t+\uptau$ where $\uptau$ is a delay in time, AF is calculated as,
\begin{equation}
R(t,t + \uptau) = \mathbb{E}[s(t)s(t+\uptau)].
\end{equation}
The average spectrum variation is relatively very-slow for a short-time scale. As a result, the occupancy state is assumed to be stationary and solely dependent on the delay $\uptau$. However on a longer scale, over a day or a week, the average behavior is not stationary and has a slowly changing average over time.
\begin{figure}
	\normalsize
	\centering
	\includegraphics[width=0.75\linewidth]{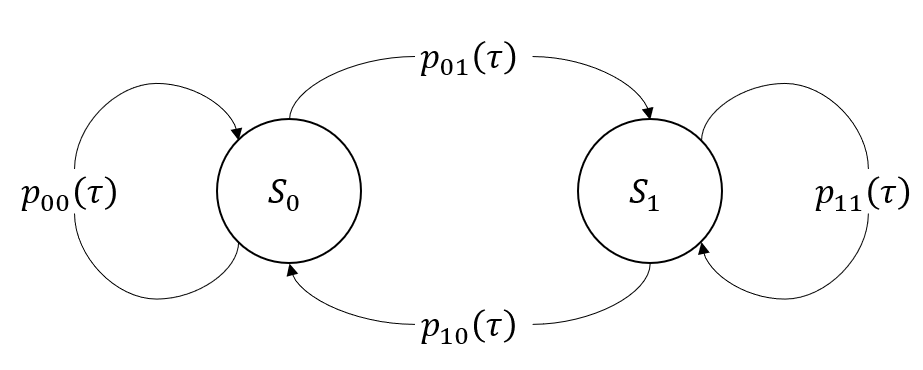}
	\caption{Channel occupancy state model.}
	\label{Fig_ChainMarkov}
\end{figure}
The expected time spent in a state $i$ before leaving to the next state $j$ is defined as the state's dwell time $\uptau_i$~\cite{leongarcia08}. We found that the autocorrelation for the captured measurements in all the sites in our campaigns can be best fit with an exponential function for the short-time scale as illustrated in Fig.~\ref{Fig_AutoCorr}, hence, the AF is simplified as follows,
\begin{equation}\label{Eqn_AutoCorr}
R(\uptau) = \mathbb{E}\left[s(0)s(\uptau)\right] = \exp\left(-\frac{\uptau}{\uptau_{\text{corr}}}\right) ,
\end{equation}
where $\uptau_{corr}$ is the average time that the channel spends in a state. Continuous-Time Markov Chain (CTMC) is a model that fits memoryless systems (exponential) and are stationary in nature (dependent on the delay only). Thus we model the spectrum as a two-state CTMC as illustrated in Fig.~\ref{Fig_ChainMarkov}. Furthermore, the transition probability for a CTMC model is ${p_{ij}(\tau) = \mathbb{P}(s(t+\uptau) = S_j|s(t) = S_i)}$ where $S_j$ is the future state and $S_i$ is the current state. The \emph{transition probability} matrix of our model is therefore constructed as follows,
\begin{equation}\label{Eqn_Hpij}
P(\uptau) =
\left[
\begin{array} {c c}
p_{00}(\uptau) &p_{01}(\uptau) \\
p_{10}(\uptau) &p_{11}(\uptau) \\
\end{array}
\right].
\end{equation}
In order to obtain the transition probabilities, we utilize the channel's initial and steady-state parameters along with the AF obtained as follows,
\begin{equation}\label{Eqn_PSolve}
p_{ij}(\uptau) = \pi_j + (p_{ij}(0) - \pi_j) R(\uptau) 	 ,
\end{equation}
where $p_{ij}(0)$ is the initial transition probability from state $i$ to $j$ and $\pi_j$ is the steady-state probability of state $j$. On the other hand, as the delay becomes sufficiently large, the future state becomes less dependent on the current state and the autocorrelation function approaches zero as the dependency on $\uptau$ becomes negligible where the probability for a channel to be in a state $j$ reaches its \emph{steady-state} value $\pi_j$. The channel duty cycle is by definition a measure that represents the busy steady-state probability whereas its complement represents the idle state steady-state probability. As a result, the steady-state probability is obtained as follows,
\begin{equation}\label{Eqn_SS}
\pi_{j} =
\begin{cases}
1-\Psi      & j = 0 \\
\Psi  		& j = 1
\end{cases} ,
\end{equation}
furthermore, by substituting the expression (\ref{Eqn_SS}) into (\ref{Eqn_PSolve}), the expression of the transition probabilities is simplified as follows,
\begin{figure}
	\normalsize
	\centering
	\includegraphics[width=\linewidth]{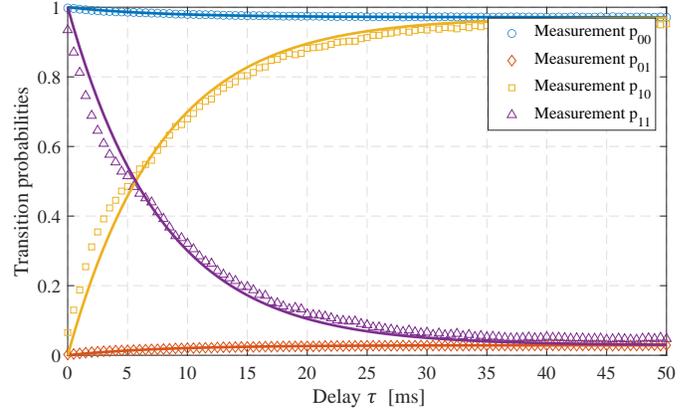}
	\caption{An example of the transition probabilities for the ISM-band with a duty cycle of ${\Psi = 0.02}$ and ${\uptau_{\text{corr}} = 7.85}$ ms for Site 1. The solid lines are calculated based on expression.~(\ref{Eqn_P}).}	\label{Fig_Transition}
\end{figure}
\begin{equation}\label{Eqn_P}
\begin{array} {l}
p_{00}(\uptau) = 1 - \Psi + \Psi R(\uptau) 	 	\\
p_{01}(\uptau) = \Psi - \Psi R(\uptau)		 	\\
p_{10}(\uptau) = 1 - \Psi - (1-\Psi)R(\uptau) 	\\
p_{11}(\uptau) = \Psi + (1-\Psi)R(\uptau) 		\\
\end{array} ,
\end{equation}
where the transition probabilities are illustrated in Fig.~\ref{Fig_Transition}. The rate of transition from state $i$ to state $j$ in a two-state CTMC model can be obtained from the transition probabilities as follows~\cite{IBE201385},
\begin{equation}\label{Eqn_Q}
q_{ij} = \left[\frac{dp_{ij}(\uptau)}{d\uptau}\right]_{\uptau=0} ,
\end{equation}
where for completeness, the two-state infinitesimal generator matrix is obtained by substituting each element in (\ref{Eqn_P}) into (\ref{Eqn_Q}) as follows,
\begin{equation}
Q = \frac{1}{\uptau_{\text{corr}}}\left[
\begin{array} {l l}
-\Psi		& \Psi 	\\
1 - \Psi    & \Psi - 1	\\
\end{array}\right].
\end{equation}
Moreover, the dwell time for state $i$ is calculated as ${\uptau_i = 1/q_{ij}}$ for $i\neq j$~\cite{IBE201385}, hence, the idle and busy dwell times are obtained as follows,
\begin{equation}
\begin{array} {l}
\uptau_0 = \uptau_{\text{corr}}/\Psi \\
\uptau_1 = \uptau_{\text{corr}}/(1 - \Psi) \\
\end{array}.
\end{equation}
\begin{figure}
	\normalsize            
	\centering
	\includegraphics[width=\linewidth]{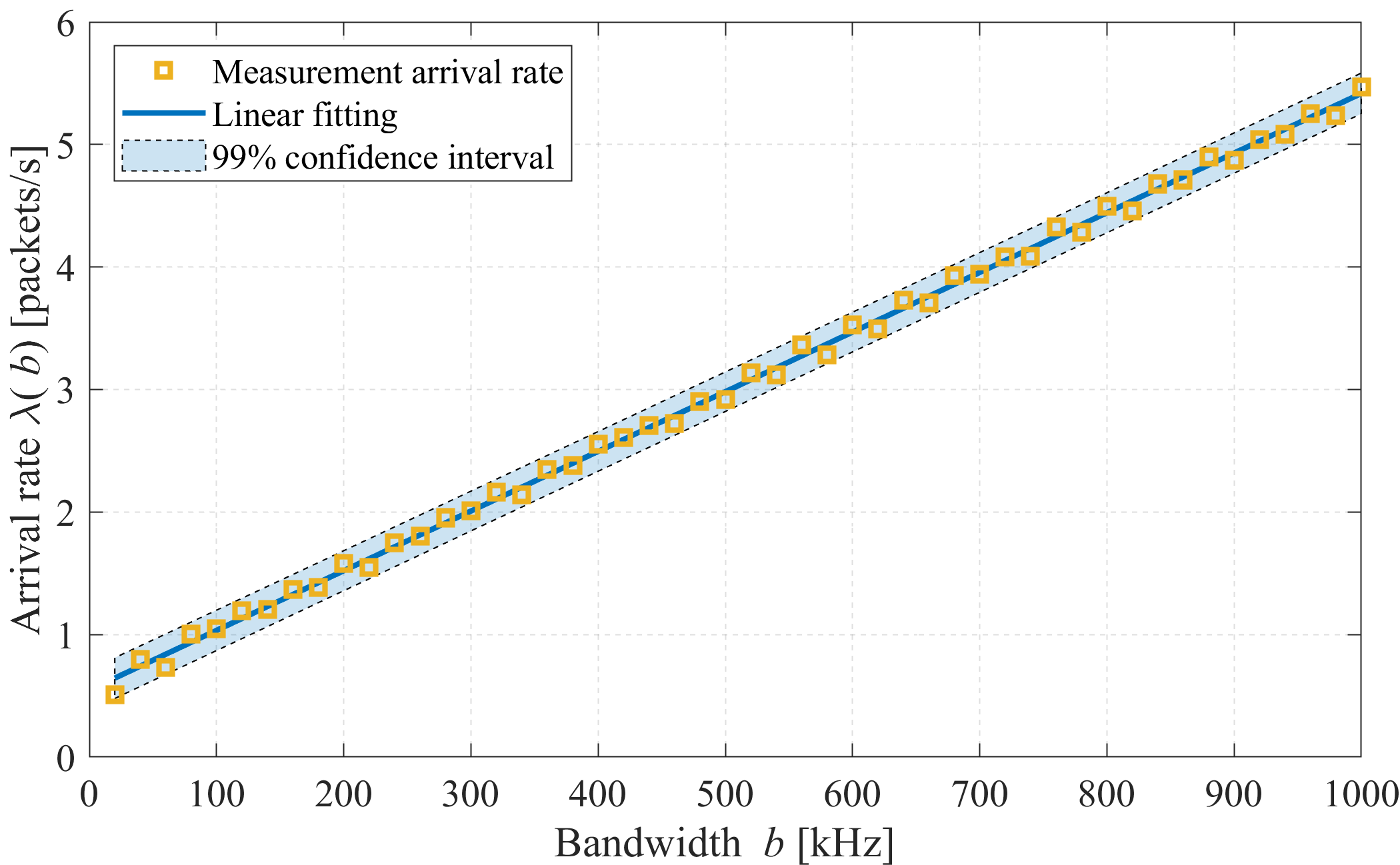}
	\caption{The discrete points represent the arrival rate of real measurements based on the clustered frames as a function of bandwidth and the solid line is based on a linear fitting  with mean squared error $3.6\times10^{-3}$.}
	\label{Fig_Arrive}
\end{figure}
\subsection{Frequency Model}
\begin{figure}
	\normalsize
	\centering
	\includegraphics[width=\linewidth]{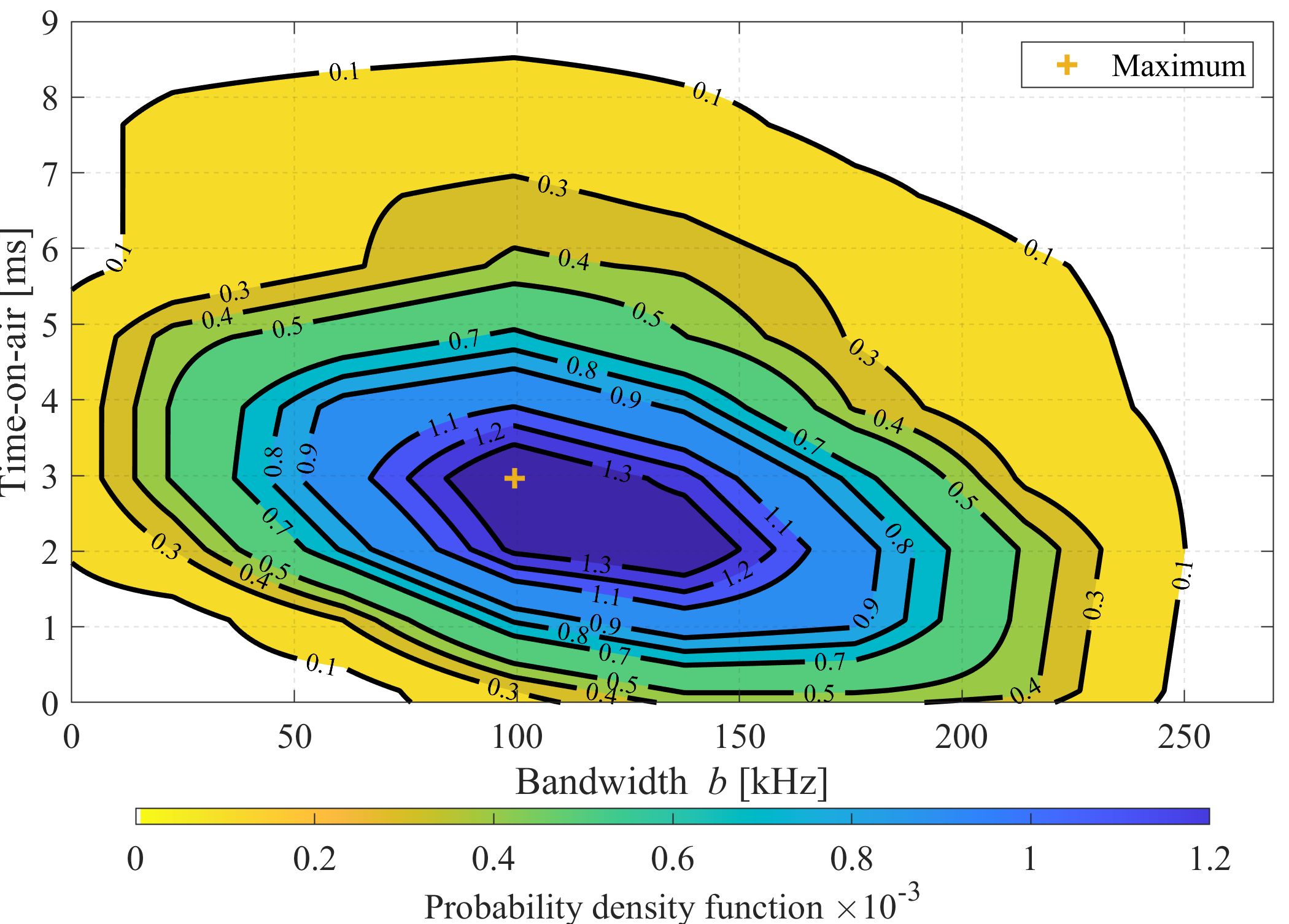}
	\caption{Joint probability density function of the bandwidth and ToA for the captured frames in the four sites combined.}
	\label{Fig_SpectPDF}
\end{figure}

The clusters obtained from (\ref{Eq_Packets}) are transmissions assumed to originate from various transmitters located at random locations operating using different modulation and coding schemes due to the nature of the ISM-band. The ISM-band regulations allow devices to transmit at any random time with any random frequency channel within the band under certain limitations to its transmit power level, duty cycle and frequency hops~\cite{ACMA_LIPD}. Note that the number of frames is a random variable that does not depend on the number of devices since they have varying duty cycles. However, the number of frames in the spectrum band is equivalent to the number of transmissions that are seen by a device observing the channel. Based on the measurements taken, the uncorrelated observed arrivals in the ISM-band can be modeled using a Poisson distribution. The arrival rate is defined as the expected number of frames per second observed by the device and is $\lambda(B) = \overline{\mathbb{N}}(C_j)/T_s$ where $\overline{\mathbb{N}}(.)$ is the average count measure and the $T_s$ is the total measurement time. If the device observing the channel has a bandwidth $b$, the expected number of frames that are observed increase linearly with the bandwidth as illustrated in Fig.~\ref{Fig_Arrive}.
The observed bandwidth is hence modeled as follows,
\begin{equation}\label{Eqn_ArrivalRate}
\lambda(b) = \frac{b}{B}\lambda(B) + \frac{B - b}{B}\lambda(0),
\end{equation}
where $\lambda(0)$ is the arrival rate at the minimum possible bandwidth of the measurement $\Delta f$ and $B$ is the ISM-band bandwidth. Furthermore, the proposed algorithm grants access to the statistics of the frames observed in the spectrum. Fig.~\ref{Fig_SpectPDF} illustrates the joint PDF of the bandwidth and ToA of the frames captured in this measurement campaign.
\subsection{Spatial Traffic Model}
Though the expected arrival model is not a direct representation of the number of devices, it is however an indicative of the number of transmissions as shown in our previous campaign in~\cite{8403749}. We introduce a normalized measure to evaluate the congestion of the spectrum $G$, where $G(b) = \lambda(b)\uptau_1$ since $\uptau_1$ is also the average time length of the packets in the channel, i.e. busy state dwell time. The normalized traffic for the channel is thus obtained as,
\begin{equation}\label{Eq_G}
G(b) = \frac{\lambda(b)\uptau_{\text{corr}}}{1-\Psi},
\end{equation}
for a given monitoring bandwidth $b$.
\begin{figure}
	\normalsize
	\centering
	\includegraphics[width=\linewidth]{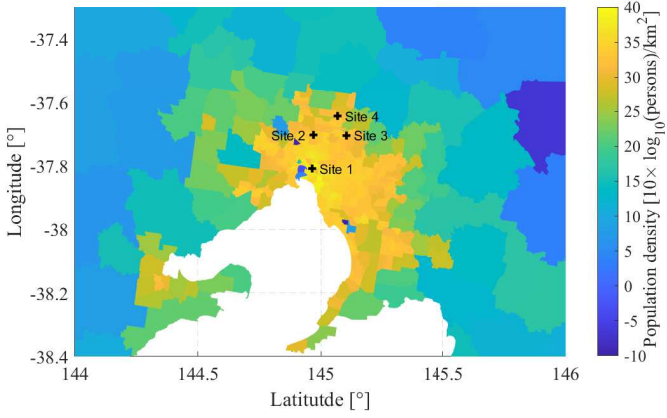}
	\caption{Population density map of Melbourne pointing the site locations of the measurement campaign based on the work related to NMSCN~\cite{NMSCP}.}
	\label{Fig_Map}
\end{figure}
\section{Experimental Setup}\label{Sec_Meas}
A measurement campaign was conducted in different sites in Melbourne with varying population densities to apply the proposed framework. Fig.~\ref{Fig_Map} illustrates the chosen site locations overlaid on top of population density map as obtained from the Australian Bureau of Statistics~\cite{Map}. We categorize the sites into three classes based on the population densities listed as follows,
\begin{itemize}
	\item Class 1 represents a \emph{dense urban} environment in Site 1 (Melbourne Central Business District) having a high mean population density of 15,000~persons/km$^2$.
	\item Class 2 represents a typical inner \emph{suburban} environment in Site 2 (Fawkner, 3060) and Site 3 (Greensborough, 3079) with average population density of 2,700 and 2,000~persons/km$^2$ respectively.
	\item Class 3 represents a typical \emph{rural} environment in Site 4 (South Morang, 3752) with a mean population density of about 1,500~persons/km$^2$.
\end{itemize}
\begin{figure}
	\normalsize
	\centering
	\includegraphics[width=0.7\linewidth]{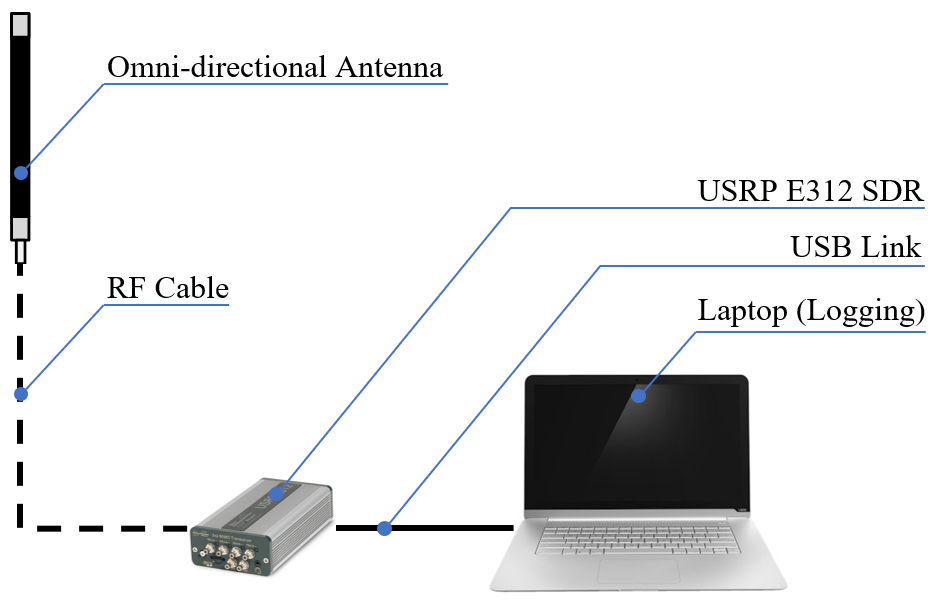}
	\caption{The setup used in the measurement campaign, using SDR E312 as a monitoring device.}
	\label{Fig_Schamtic}
\end{figure}
\begin{table}[t]
	\caption{Measurement Setup Parameters}
	\centering
	\begin{tabular}{ l l l }
		\hline\hline
		Parameter 					 				& Value 	&Unit	\\
		%heading
		\hline
		Frequency band (915~MHz	ISM-band~\cite{ACMA_LIPD}) & 915 - 928 & MHz   \\ 
		Antenna gain @921~MHz~\cite{Antenna915}		& 6			& dBi 	\\ 
		SDR sampling rate $f_\text{s}$ 		    			& 30		& MHz   \\
		SDR pre-amplifier gain (manual mode$^\dagger$)	& 70~\cite{USRP}		& dB 	\\
		SDR measured noise floor       				& -154		& dBm/Hz\\ 
		SDR measured noise figure @70 dB gain		& 17		& dB	\\
		Selected detection threshold				& -140 		& dBm/Hz \\
		\hline
	\end{tabular}
	\begin{tabularx}{\textwidth}{l}
		$^{\dagger}$ The AGC is turned off in the sensing mode to maintain noise floor \\ level.
	\end{tabularx}
	\label{Table_Setup}
\end{table}
\begin{table}[t]
	\caption{Data Capturing and STFT Parameters}
	\centering
	\begin{tabular}{ l l l l }
		\hline\hline
		Parameter 						& Symbol 		& Value 			&Unit 		\\
		%heading
		\hline
		Window size	(Hamming)			&$N_\text{w}$	& $15\times 10^3$	&samples	\\
		Number of DFT points			&$N_\text{f}$	& $15\times 10^2$	&samples	\\
		Time resolution				 	&$\Delta t$		& 0.5				&ms    	    \\
		Averaged frequency points		&$z$			& 10				&samples	\\
		Frequency resolution		    &$\Delta f$ 	& 20				&kHz       	\\ 
		Windowing overlap			    &-				& 0					&\%		   	\\
		\hline
	\end{tabular}
	\label{Table_Data}
\end{table}
We utilize a USRP E312~\cite{USRP} SDR in our measurement campaigns due to its low average noise floor and front-end filters which are suitable for outdoor sensing. The measurement setup consists of a 921~MHz monopole antenna connected to the SDR via an RF cable. The SDR is connected via a USB cable to the logging laptop that is controlling the SDR using MATLAB, a schematic of which is illustrated in Fig.~\ref{Fig_Schamtic}. Table~\ref{Table_Setup} summarizes the parameters of the measurement setup. The SDR is configured to capture IQ-samples in the ISM-band (915 - 928~MHz in Australia~\cite{ACMA_LIPD}). The logging computer runs an automated MATLAB script that processes the signal captured. The captured samples are windowed into smaller strides of size $N_w$ processed using a discrete Short-Time Fourier Transform (STFT) with a Hamming window. The resulting time resolution of the STFT is $\Delta t = N_\text{w}/ f_\text{s}$. In order to meet the Nyquist criteria, the frequency resolution of the spectrogram obtained is set to $1/\Delta t$. We average the frequency points by a factor $z$ to reduce the size of the needed storage and enhance the capturing performance resulting in $\Delta f = z/\Delta t$ where the number of DFT points $N_\text{f} = f_\text{s}/\Delta f$. Table~\ref{Table_Data} summarizes the utilized STFT parameters in the script. The spectrogram matrix is obtained by calculating the normalized squared magnitude of the STFT as follows,
\begin{equation}
X[n,k] = \left |\frac{1}{N_\text{w}}\sum_{m=-N_\text{w}/2}^{N_\text{w}/2}x[m]w[n-m]\exp\left(\frac{-j2\pi kn}{N_\text{f}}\right)\right|^2,
\end{equation}
where $x[n]$ is the windowed captured samples, $w[n]$ is the hamming window, $m$ is the stride number, $n$ is the discrete time variable, and $k$ is the discrete frequency variable.
%The resulting spectrogram matrix is therefore ${A \times N_f}$ in size where $A=\floor{N_s/N_w}$.
The spectrogram matrix has a range of ${-f_s/2\leq k\Delta f\leq f_s/2}$ and ${0\leq n\leq (N_s-1)/f_s}$ in the frequency and time domains respectively with no overlap between the window hops.

The normalized traffic obtained from the measurements is summarized in Table~\ref{Table_Results} for the four different sites. This measure can be utilized by devices to change their transmission frequency if the normalized traffic is high in the current operation sub-band.
\begin{table}[t]
	\caption{Measurement Campaign Results}
	\centering
	\begin{tabular}{l l l l l}
		\hline\hline
		Parameter &$\Psi$ [\%]	&$\uptau_\text{corr}$ [ms]	&$\lambda(B)$ [packets/s] & $G(B)$ 	\\
		%heading
		\hline
		Site 1	& 0.35	& 7.07   	& 57.86 	& 0.41\\ 
		Site 2  & 0.17	& 6.93		& 28.28 	& 0.20\\ 
		Site 3 	& 0.06	& 9.68		& 6.87 		& 0.07\\
		Site 4	& 0.05 	& 6.27		& 12.42  	& 0.08\\
		\hline
	\end{tabular}
	\label{Table_Results}
\end{table}
\section{Conclusion and Future Work}\label{Sec_Conc}
This paper presented a novel unsupervised algorithm that estimates the actual occupancy state from the observed state of the ISM-band unlicensed channel. The algorithm enhances energy detection leading to a lower probability of false alarm and misdetection. Moreover, the channel temporal and frequency behavior were modeled based on a continuous-time Markov chain and Poisson arrival rate respectively. Measurement campaigns were conducted in four different environments to better understand the spatial effect on the channel. The paper concludes that the IoT spectrum can be modeled using one measure referred to as the normalized traffic which is dependent on the underlying urban environments. The models presented assist in identifying statistics such as the average busy time and the average number of interfering transmissions in a channel. These statistics enable IoT devices to have prior or continuous knowledge of the channel depending on the deployment. The framework can be deployed in an online fashion where the spectrum data are saved in blocks. The data blocks can be fixed or dynamic in size limited by the SDR buffer size. The algorithm utilizes each block to extract the occupancy since it does not require all the data to be processed in one batch. Future work includes modeling the power level which provides a third degree of freedom in terms of the extracted statistics of the transmissions.

% Detail changes (before and after)

\ifCLASSOPTIONcaptionsoff
\newpage
\fi
\bibliographystyle{IEEEtran}
\bibliography{TemporalSpectrumIoT}

\end{document}